\documentclass[conference]{IEEEtran}
\usepackage{cite}
\usepackage{amsmath,amssymb,amsfonts}
\usepackage{algorithmic}
\usepackage{graphicx}
\usepackage{textcomp}
\usepackage{xcolor}

\usepackage{bm}
\usepackage{booktabs}
\usepackage{multirow}
\usepackage{subcaption}
\usepackage{hyperref}
\usepackage{cleveref}
\usepackage{threeparttable}
\def\BibTeX{{\rm B\kern-.05em{\sc i\kern-.025em b}\kern-.08em
    T\kern-.1667em\lower.7ex\hbox{E}\kern-.125emX}}

\begin{document}

\title{Heterogeneous Graph Tree Networks}

\author{\IEEEauthorblockN{Nan Wu}
\IEEEauthorblockA{\textit{Institute for Financial Services Analytics} \\
\textit{University of Delaware}\\
Newark, USA \\
nanw@udel.edu}
\and
\IEEEauthorblockN{Chaofan Wang}
\IEEEauthorblockA{
San Jose, USA \\
chaofanwang123@gmail.com}
}


\maketitle

\begin{abstract}
Heterogeneous graph neural networks (HGNNs) have attracted increasing research interest in recent three years. Most existing HGNNs fall into two classes. One class is meta-path-based HGNNs which either require domain knowledge to handcraft meta-paths or consume huge amount of time and memory to automatically construct meta-paths. The other class does not rely on meta-path construction. It takes homogeneous convolutional graph neural networks (Conv-GNNs) as backbones and extend them to heterogeneous graphs by introducing node-type- and edge-type-dependent parameters. Regardless of the meta-path dependency, most existing HGNNs employ shallow Conv-GNNs such as GCN and GAT to aggregate neighborhood information, and may have limited capability to capture information from high-order neighborhood. In this work, we propose two heterogeneous graph tree network models: Heterogeneous Graph Tree Convolutional Network (HetGTCN) and Heterogeneous Graph Tree Attention Network (HetGTAN), which do not rely on meta-paths to encode heterogeneity in both node features and graph structure. Extensive experiments on three real-world heterogeneous graph data demonstrate that the proposed HetGTCN and HetGTAN are efficient and consistently outperform all state-of-the-art HGNN baselines on semi-supervised node classification tasks, and can go deep without compromising performance.
\end{abstract}


\section{Introduction}
\label{intro}
Although Graph Neural Network (GNN) has demonstrated its success in many graph learning applications such as link predictions \cite{zhang2018link, zhu2021neural}, fraud detection \cite{wang2019semi, dou2020enhancing}, personalized recommendations \cite{yin2019deeper, fan2019metapath, fan2019graph}, and pattern recognition \cite{shi2020point, shen2018person}, most existing GNNs are only applicable to homogeneous graphs with single type of nodes and edges. As a matter of fact, a large portion of real-world graphs contain various types of nodes and/or edges, which are known as \textit{heterogeneous graphs}. Attributes associated with different node types usually lie in different feature spaces. For example, a citation graph may contain three types of nodes: authors, papers and venues; and four types of edges: author-paper, paper-venue and their reverse connections. The complex structure and rich semantic information make it nontrivial to extend applications of a homogeneous GNN to heterogeneous graphs.

Heterogeneous graph neural networks (HGNNs) have attracted increasing research interest in recent three years, yet are still under explored. Most existing HGNNs fall into two classes. One class is meta-path-based HGNNs which either require domain knowledge to handcraft meta-paths \cite{wang2019heterogeneous, fu2020magnn} or consume huge amount of time and memory to automatically construct meta-paths \cite{yun2019graph}. The other class does not rely on meta-path construction \cite{schlichtkrull2018modeling, hu2020heterogeneous, lv2021we}. This class of models takes homogeneous convolutional graph neural networks (Conv-GNNs) as backbones and extend them to heterogeneous graphs by introducing node-type- and edge-type-dependent parameters. Regardless of the meta-path dependency, most existing HGNNs employ shallow Conv-GNNs such as GCN \cite{kipf2017GCN} and GAT \cite{VelickovicCCRLB18} to aggregate neighborhood information, and therefore may have limited capability to capture information from high-order neighborhood.

\cite{wu2022gtnet} recently proposes two deep convolutional graph tree networks which adopt a different message passing scheme from GCN and GAT, and demonstrates the deep capability of the proposed models both theoretically and experimentally. In light of the superior performance of the two graph tree network models proposed by \cite{wu2022gtnet}, we propose two heterogeneous graph tree network models: Heterogeneous Graph Tree Convolutional Network (HetGTCN) and Heterogeneous Graph Tree Attention Network (HetGTAN). Both models include three modules: (1) a node-type-specific transformation to project node features into the same vector space; (2) an edge-type-specific graph tree convolutional network (GTCN) \cite{wu2022gtnet} or graph tree attention network (GTAN) \cite{wu2022gtnet} layer to aggregate one-hop neighbors connected via the same type of edges; (3) a target-specific aggregator to combine representations of the target node from different edge types. Our proposed models do not depend on meta-paths to encode heterogeneity in both node features and graph structure, and can go deep to incorporate information from high-order neighborhood without compromising performance.

When evaluating our models, we find that all popular HGNNs underestimate the performance of the two most popular baseline GNNs - GCN and GAT by simply ignoring the heterogeneity of nodes and edges \cite{yun2019graph, hu2020heterogeneous, lv2021we} or testing on meta-path based homogeneous subgraphs and reporting the best performance \cite{wang2019heterogeneous, fu2020magnn}. To better benchmark HGNNs, we propose HetGCN and HetGAT to extend the vanilla GCN and GAT to heterogeneous graphs. With the same architecture of HetGTCN and HetGTAN, HetGCN and HetGAT also include three modules: (1) a node-type-specific transformation to project node features into the same vector space; (2) an edge-type-specific GCN or GAT layer to aggregate one-hop neighbors connected via the same type of edges; (3) a target-specific aggregator to combine representations of the target node from different edge types.

We conduct comprehensive experiments on three real-world heterogeneous graph datasets on semi-supervised node classification tasks, and make fair comparisons of the proposed HetGTCN and HetGTAN with nine baseline HGNN models including HetGCN and HetGAT. We find that HetGAT already outperforms or matches the other baseline HGNNs. Our proposed HetGTCN and HetGTAN consistently and significantly outperform all baseline models. We also demonstrate the efficiency of our proposed HetGTCN and HetGTAN by comparing the runtime and memory consumption with the other baselines. Moreover, our proposed HetGTCN and HetGTAN models can go deep without compromising performance, while the performance of most baseline HGNNs degrades significantly with an increasing number of model layers.

The major contributions of this work are summarized as follows:
\begin{enumerate}
	\item[(1)] We propose two efficient and effective heterogeneous graph neural network models: HetGTCN and HetGTAN which can encode heterogeneity in both node features and graph structure without designing meta-paths. Three different aggregator functions are explored in this work: semantic attention aggregator, simple mean aggregator, and weighted sum aggregator with learnable weights. 
	\item[(2)] We conduct extensive experiments and demonstrate that the proposed HetGTCN and HetGTAN outperform all baseline HGNNs, and can go deep without compromising performance.
	\item[(3)]We propose two baseline HGNNs - HetGCN and HetGAT which extend the two most popular homogeneous GNNs - GCN and GAT to heterogeneous graphs. We experimentally demonstrate that HetGAT outperforms or matches the other baseline HGNNs.
	\item[(3)] We provide unified pre-processing on three real-world heterogeneous graph datasets to promote fair comparisons between the proposed models and baselines.
\end{enumerate}

\section{Related Work}
\label{literature}
\textbf{Convolutional Graph Neural Networks (Conv-GNNs).} The class of spatial-based Conv-GNNs has attracted considerable attention over the past few years following the publication of GCN \cite{kipf2017GCN}. These models adopt a message passing scheme to aggregate neighborhood information, among which the vanilla GCN \cite{kipf2017GCN} and GAT \cite{VelickovicCCRLB18} are arguably the two most popular baseline GNNs. GCN uses a normalized mean aggregator to aggregate neighborhood information, while GAT uses an additive attention \cite{bahdanau2014neural} aggregator to aggregate neighborhood information. One common issue with GCN and GAT is the over-smoothing problem \cite{li2018deeper, liu2020towards, oono2019graph, chen2020measuring, wang2019improving} that the performance of GCN and GAT degrades significantly with an increasing number of model layers. Therefore, both GCN and GAT are often referred to as shallow models.

\textbf{Heterogeneous Graph Neural Networks (HGNNs).} Witnessing the success of homogeneous GNNs in different domains on graph-structured data, many efforts have been devoted to developing GNNs for heterogeneous graphs in recent three years. Most existing HGNNs take a homogeneous Conv-GNN such as GCN \cite{kipf2017GCN} and GAT \cite{VelickovicCCRLB18} as their backbone, and encode the graph heterogeneity by either meta-paths or node-type- and edge-type-dependent parameters. They are essentially a combination of meta-path-specific or edge-type-specific Conv-GNNs, and can be described by a general heterogeneous message passing neural network (MPNN) formulated in \Cref{eqn:mpnn} which is an extension of \cite{gilmer2017neural}.
\begin{align}
	\label{eqn:mpnn}
	\begin{split}
		&\bm{m}_u^{l+1} = \sum_{k=1}^{K_a} {\theta_{k,a}^l \sum_{v\in \mathcal{N}_u^{k}} M_{k}^l\left(\bm{h}_u^l, \bm{h}_{v}^l, \bm{h}^l_{k(u,v)}\right) } \\
		&\bm{h}_u^{l+1} = U_l \left(\bm{h}_u^l, \bm{m}_u^{l+1}\right)
	\end{split}
\end{align}
where $a$ is the node type of the target node $u$, and $\bm{h}_u$ is the hidden feature of node $u$. $k$ is the relation type, which can be referred to as either meta-path type or edge type. $K_a$ is the number of relation types connected to type $a$ nodes, $\mathcal{N}_u^k$ is the set of relation-type-specific neighbors of node $u$, and $\bm{h}_{k(u,v)}$ is the hidden feature of meta-path instance $k(u,v)$ or edge $(u,v)$. $M_k(\cdot)$ is a relation-type-specific message passing function, and $U(\cdot)$ is a node feature update function \cite{gilmer2017neural}. $\theta_{k,a}$ denotes the importance of different relation types for target nodes of type $a$, which can be either fixed or learnable. $l$ denotes the message passing layer number.

Representative meta-path-based HGNNs include HAN \cite{wang2019heterogeneous}, MAGNN \cite{fu2020magnn}, and GTN \cite{yun2019graph}. Among which both HAN and MAGNN can fit into the general MPNN framework described in \Cref{eqn:mpnn}. HAN \cite{wang2019heterogeneous} is one of the earliest attempts on HGNNs with $M_k^l(\cdot) = \text{GAT}_k^l(\bm{h}_u^l, \bm{h}_{v}^l)$, $U_l(\cdot) = \bm{m}_u^{l+1}$, and $\theta_{k,a}^l$ is a learnable semantic attention weight. It takes GAT as its backbone to aggregate meta-path-based neighbors, and a semantic attention aggregator to combine representations of different meta-paths. HAN ignores all intermediate nodes along a meta-path instance. MAGNN \cite{fu2020magnn} improves over HAN by introducing a meta-path instance encoder to include all nodes along a meta-path instance, where  $M_k^l(\cdot) = \text{GAT}_k^l(\bm{h}_u^l, \bm{h}^l_{k(u,v)})$. Both HAN and MAGNN rely on domain knowledge to handcraft meta-paths for different downstream tasks, and therefore may suffer from information loss. GTN \cite{yun2019graph} learns a new graph structure defined by a weighted sum of all meta-path-based adjacency matrices, and then applies GCN to the new graph structure. Although GTN is able to extract valuable meta-paths automatically, it requires huge amount of time and memory to learn the new graph which is constructed by multiplication of soft-selected adjacency matrices of each edge type. One common issue faced by meta-path-based HGNNs is that they are not scalable when high-order meta-paths are desired to capture information from high-order neighborhood because the number of high-order meta-path instances may increase dramatically. Take the DBLP dataset \cite{Fey/Lenssen/2019} as an example, the number of edges with type "Author-Paper" (AP) and "Paper-Conference" (PC) is 19,645 and 14,328 respectively, while the number of meta-path "APCPA" is 5,000,495. 

Representative meta-path-free HGNNs include RGCN \cite{schlichtkrull2018modeling}, HGT \cite{hu2020heterogeneous}, and SimpleHGN \cite{lv2021we}, which all can fit into the general MPNN framework described in \Cref{eqn:mpnn}. RGCN \cite{schlichtkrull2018modeling} is proposed to learn from multi-relational knowledge graphs with a single node type. It takes GCN as its backbone, and can be regarded as a weighted sum of edge-type-specific GCNs, where $M_k^l(\cdot) = \text{GCN}_k^l(\bm{h}_u^l,\bm{h}_{v}^l)$, and $U_l(\cdot) = \bm{m}_u^{l+1}$. HGT \cite{hu2020heterogeneous} is a transformer-based HGNN, which uses transformer attention as $M_k$, and $\theta_{k,a} = 1$. Skip connection is applied with $U_l(\cdot) = \sigma(\bm{m}_u^{l+1})\bm{W}_a + \bm{h}_u^l$. It extends \cite{li2019graph} to heterogeneous graphs by introducing node-type-dependent transformation matrices for query, key, and value projection, respectively, and edge-type-dependent weight matrices for transformer attention and message transformation, respectively. SimpleHGN \cite{lv2021we} extends GAT to heterogeneous graphs by introducing a learnable edge type embedding and a corresponding edge-type-dependent transformation matrix in the node pair attention. It uses $\text{GAT}$ as $M_k$ and $\theta_{k,a} = 1$. It also applies the skip connection with $U_l(\cdot) = \bm{m}_u^{l+1} + \bm{h}_u^l$.

DMGI \cite{park2020unsupervised} is a representative unsupervised HGNN based on Deep Graph Infomax (DGI) \cite{velickovic2019deep}. It is trained to maximize the mutual information between the node embedding and the graph embedding regarding each relation type, and minimize a consensus regularizer, where the relation-type-specific node embeddings are generated by a relation-type-specific GCN layer. The relation can either refer to an edge or a meta-path instance.

Although the aforementioned HGNNs have been successfully applied to many applications on heterogeneous graphs, none of them has discussed about their deep capability. We demonstrate in Section \ref{exp:deep} that all these HGNNs have limited capability to capture information from high-order neighborhood and show performance decay with an increasing number of model layers.


\section{Preliminary}
In this section, we introduce the definition of heterogeneous graphs and notations used throughout this paper. We then review the two homogeneous graph tree network models proposed in \cite{wu2022gtnet}.

\subsection{Heterogeneous Graph}
A heterogeneous graph is defined as $\mathcal{G} = (\mathcal{V}, \mathcal{E}, \mathcal{T}^v, \mathcal{T}^e)$ with a node type mapping function $\phi$: $\mathcal{V} \to \mathcal{T}^v$ and an edge type mapping function $\psi$: $\mathcal{E} \to \mathcal{T}^e$, where $\mathcal{V}$ is the set of nodes, $\mathcal{E}$ is the set of edges, $\mathcal{T}^v$ is the set of node types, and $\mathcal{T}^e$ is the set of edge types. $|\mathcal{T}^v|$ and $|\mathcal{T}^e|$ denote the number of node types and edge types respectively, where $|\mathcal{T}^v| + |\mathcal{T}^e| > 2$. A heterogeneous graph can be fully described by a set of adjacency matrices $\{\bm{A}_k \}_{k=1}^{|\mathcal{T}^e|}$ and the corresponding set of degree matrices $\{\bm{D}_k \}_{k=1}^{|\mathcal{T}^e|}$, where $\bm{A}_k$ is the adjacency matrix of edge type $k$. Let $\mathcal{V}_a$ denote the set of nodes of type $a$. Each node $u \in \mathcal{V}_a$ is associated with an input feature vector $\bm{x}_u \in \mathbb{R}^{1 \times d_a}$. The input feature map for all nodes of type $a$ is denoted as $\bm{X}_a \in \mathbb{R}^{|\mathcal{V}_a| \times d_a}$. Let $\bm{h}_u^l \in \mathbb{R}^{1\times f}$ denote the hidden feature of node $u$ at the $l^\text{th}$ layer, and $\mathcal{N}_u^k$ denote the set of one-hop neighbors of node $u$ connected via type $k$ edges.

\subsection{Graph Tree Networks}
Trees can intuitively represent the complex structure of a graph, where each target node and its neighborhood form a tree with the target node being the root node and its neighbors being the subnodes. The depth of the graph tree is determined by the size of desired receptive field.

\cite{wu2022gtnet} assumes that each node preserves its initial information prior to receiving new information from its child nodes in the graph tree, and proposes Graph Tree Networks (GTNets) with a message passing scheme following this assumption. In GTNet, the representation of a target node is updated by aggregating its neighbors' updated hidden features from the previous layer and its own \textbf{initial feature}. In contrast, GCN and GAT update a target node's representation by aggregating its neighbors' and its own updated hidden features from the previous layer. Following the architecture of GTNet, \cite{wu2022gtnet} proposes two homogeneous graph tree network models - Graph Tree Convolutional Network (GTCN) and Graph Tree Attention Network (GTAN).

The message passing rule in one GTCN layer is described as \cite{wu2022gtnet}
\begin{equation}
	\label{eqn:gtcn_node}
	\bm{h}_u^l = \sum\nolimits_{v\in \mathcal{N}_u}{\hat{\bm{A}}_{uv} \bm{h}_v^{l+1}} + \hat{\bm{A}}_{uu} \bm{z}_u
\end{equation}
where $\hat{\bm{A}} = \tilde{\bm{D}}^{-\frac{1}{2}} \tilde{\bm{A}} \tilde{\bm{D}}^{-\frac{1}{2}}$ is the symmetrically normalized adjacency matrix. For a directed graph, the adjacency matrix $\bm{A}$ is asymmetric, $\tilde{\bm{A}}$ can be normalized by its inverse degree matrix $\tilde{\bm{D}}^{-1}$, i.e. $\hat{\bm{A}} = \tilde{\bm{D}}^{-1} \tilde{\bm{A}}$. $\tilde{\bm{A}} = \left(\bm{A} + \bm{I}\right)$ is the adjacency matrix with added self-loops. $\tilde{\bm{D}}$ is the degree matrix of $\tilde{\bm{A}}$, where $\tilde{\bm{D}}_{uu} = \sum\nolimits_v \tilde{\bm{A}}_{uv}$. $l$ denotes the order (i.e., hop) of neighborhood to the target node, which is in an reverse order of the model layer number. $\bm{z}_u$ is the initial hidden feature of node $u$ such that $\bm{h}_u^L = \bm{z}_u = \text{MLP}(\bm{x}_u)$, where $L$ is the model depth.

The message passing rule in one GTAN layer is described as \cite{wu2022gtnet}
\begin{equation}
	\label{eqn:gtan_node}
	\bm{h}_u^l = \text{ELU}\left( \sum\nolimits_{v\in \mathcal{N}_u} \alpha_{u,v}^{l+1} \bm{h}_v^{l+1} + \alpha_{u,u}^{l+1} \bm{z}_u \right)
\end{equation}
where $\alpha_{u,v}$ is the attention weight for the node pair $(u,v)$ calculated by the additive attention \cite{bahdanau2014neural} as
\begin{equation}
	\label{eqn:attention_weight}
	\alpha_{u,v}^l =
	\begin{cases}
		\text{softmax} \left( \text{LeakyReLU} \left( \left[\bm{z}_u \parallel \bm{h}_v^l\right] \bm{a}^l \right)\right) & \text{if $u \neq v$}\\
		\text{softmax} \left(\text{LeakyReLU} \left( \left[\bm{z}_u \parallel \bm{z}_u\right] \bm{a}^l \right) \right) & \text{if $u = v$}
	\end{cases}
\end{equation}
where $\bm{a}^l$ is a learnable attention vector shared by all edges.

Built on GTCN and GTAN, we propose two heterogeneous graph tree network models - Heterogeneous Graph Tree Convolutional Network (HetGTCN) and Heterogeneous Graph Tree Attention Network (HetGTAN) in Section \ref{method}.

%
%
%

\section{Heterogeneous Graph Tree Networks}
\label{method}
\begin{figure}[htpb]
	\centering
	\includegraphics[width=\linewidth]{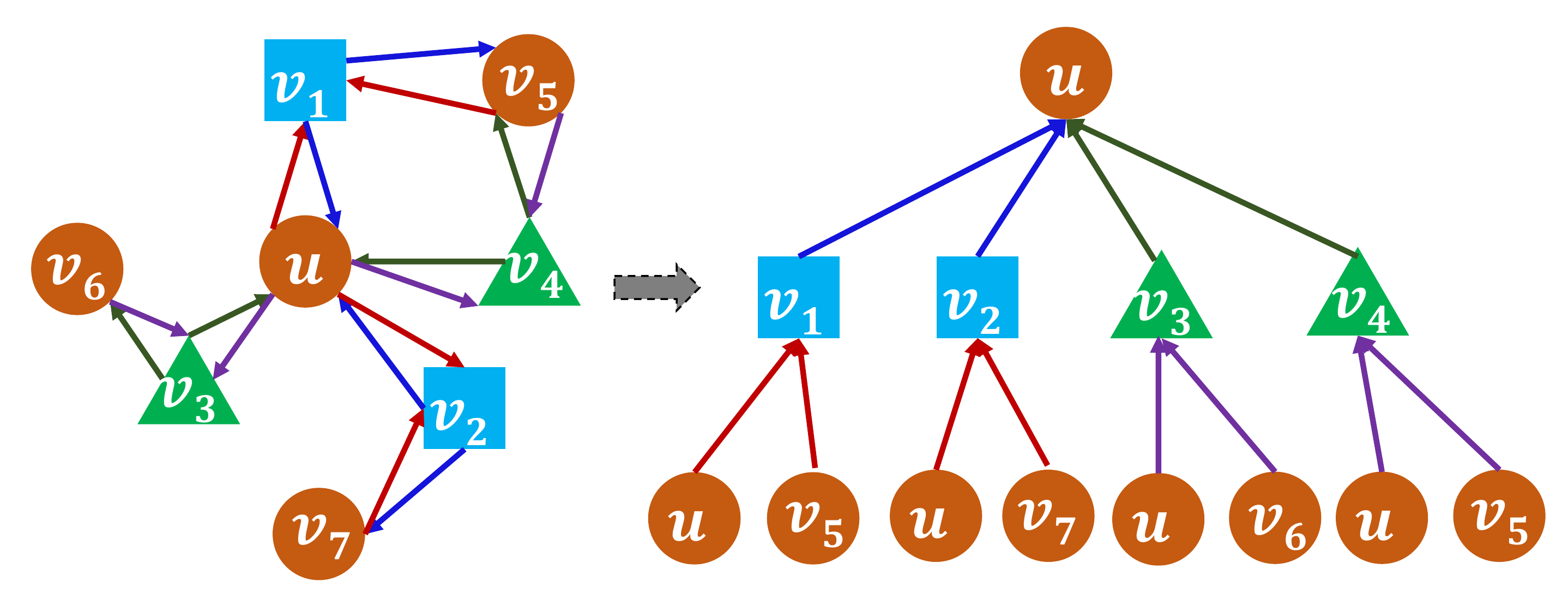}
	\caption{Tree representation of a target node $u$ and its two-hop neighborhood in a sample heterogeneous graph.}
	\label{fig:tree}
\end{figure}

\begin{figure*}[!htbp]
	\centering
	\includegraphics[width=\linewidth]{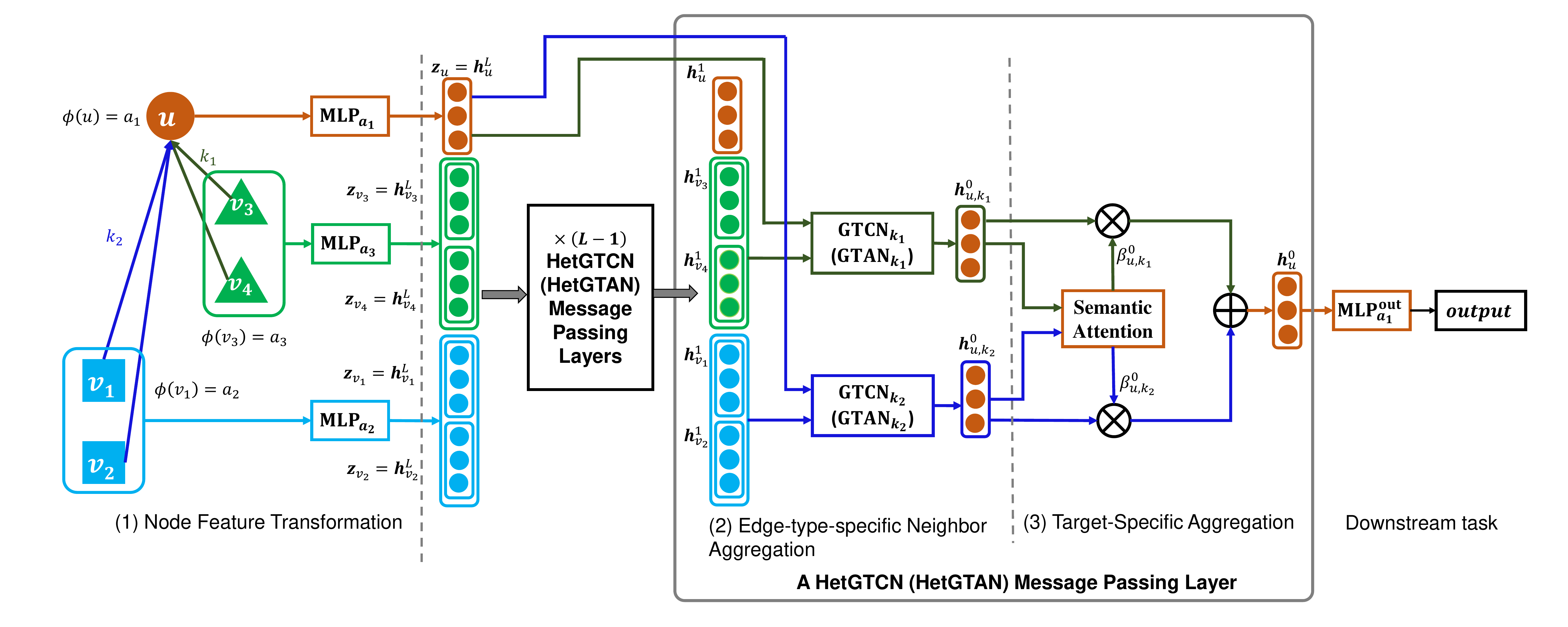}
	\caption{The overall architecture of HetGTCN and HetGTAN. Each model is consisted of three modules: (1) a node-type-specific transformation to project node features into the same vector space; (2) an edge-type-specific GTCN or GTAN layer to aggregate neighbors with the same type of relations; (3) a target-specific aggregator to combine representations from different edge types. For simplicity and clarity, we show the pipeline for a sub-graph with node $u$ and its one-hop neighbors. Node types, edge types and their dependency are denoted with different colors.}
	\label{fig:architecture}
\end{figure*}

Figure \ref{fig:tree} shows an example of a heterogeneous graph and its tree representation of node $u$ and its two-hop neighborhood. There are three types of nodes, and four types of edges in the sample graph. We use different colors to distinguish the node types and edge types. For a heterogeneous graph, each path from a subnode to the root node in its tree representation represents a meta-path instance. Messages propagate upward along each meta-path instance to pass information from subnodes to the target node. A parent node may have multiple types of child nodes connected via different types of edge. Nodes of different types may be associated with features lying in different feature spaces. An effective heterogeneous graph tree model should be able to encode such heterogeneity in both node contents and graph structure.

In this section, we propose two heterogeneous graph tree network models: Heterogeneous Graph Tree Convolutional Network (HetGTCN) and Heterogeneous Graph Tree Attention Network (HetGTAN). Both models are consisted of three modules: (1) a node-type-specific transformation to project node features into the same vector space; (2) an edge-type-specific GTCN \cite{wu2022gtnet} or GTAN \cite{wu2022gtnet} layer to aggregate one-hop neighbors connected via the same type of edges; (3) a target-specific aggregator to combine representations of the target node from different edge types. Three different aggregator functions are explored in Section \ref{ablation} which are semantic attention, simple mean, and weighted sum with learnable weights. \Cref{fig:architecture} shows the overall architecture of the proposed HetGTCN and HetGTAN with a semantic attention aggregator. We use different colors to denote types of nodes and edges, and the type dependency of each component. This architecture can turn any homogeneous GNN into a heterogeneous GNN by replacing the second module with the particular edge-type-specific homogeneous GNN.

We introduce each module of HetGTCN and HetGTAN step by step in the following sections.

\subsection{Heterogeneous Graph Tree Convolutional Network (HetGTCN)}
\label{HetGTCN}
We assume the order of neighborhood under consideration is $L$, i.e., the model depth is $L$.

\textbf{Node Feature Transformation.} In heterogeneous graphs, attributes associated with different node types usually lie in different feature spaces. To aggregate information from different types of nodes, we need to project node features into the same vector space. We first apply a node-type-specific MLP to transform all node features into the same vector space. For a node $u \in \mathcal{V}_a$, i.e. $\phi \left( u \right) = a$, we have
\begin{equation}
	\bm{z}_u = \sigma \left( \bm{x}_u \bm{W}_{0,a} + \bm{b}_{0,a} \right)
\end{equation}
and 
\begin{equation}
	\bm{h}_u^L = \bm{z}_u
\end{equation}
where $\bm{x}_u \in \mathbb{R}^{1 \times d_a}$ is the input feature vector of node $u$, $\bm{z}_u \in \mathbb{R}^{1\times f}$ is the projected input feature vector of node $u$, which is also the initial hidden feature $\bm{h}_u^L$ for the first message passing layer. $\bm{W}_{0,a} \in \mathbb{R}^{d_a \times f}$ is the learnable weight matrix for node type $a$, and $\bm{b}_{0,a} \in \mathbb{R}^{1\times f}$ is the learnable bias vector for node type $a$.

\textbf{Edge-type-specific Neighbor Aggregation.} After all node features are transformed into the same vector space, we apply an edge-type-specific GTCN layer to aggregate information from neighbors connected to the target node via the same type of edges.

\begin{equation}
	\bm{h}^{l}_{u,k} = \sum\nolimits_{v\in \mathcal{N}_u^k}{\hat{\bm{A}}_{k,uv} \bm{h}_v^{l+1}} + \hat{\bm{A}}_{k,uu} \bm{z}_u
\end{equation}
where $\hat{\bm{A}}_k = \tilde{\bm{D}}_k^{-1} \tilde{\bm{A}}_k$, which is the normalized adjacency matrix of edge type $k$.

Assume there are $K_a$ different types of edges connected to the node type $a$, the edge-type-specific neighbor aggregation generates a set of $K_a$ edge-type-specific representations for a target node $u \in \mathcal{V}_a$, denoted as $\left\{\bm{h}^l_{u,1}, \bm{h}^l_{u,2}, \ldots, \bm{h}^l_{u,K_a} \right\}$.

\textbf{Target-Specific Aggregation.} Various aggregators can be employed to combine the edge-type-specific representations for a target node. Here, we leverage the semantic attention mechanism \cite{lin2017structured} to learn the importance of each edge type to a target node, and aggregate the edge-type-specific representations to obtain the final representation of the target node. Another two aggregators are investigated in Section \ref{ablation}. 

For $u \in \mathcal{V}_a$, we have:
\begin{equation}
	\bm{h}_u^l = \sum_{k=1}^{K_a} {\beta_{k}^l \cdot \bm{h}_{u,k}^l}
\end{equation}
where $\beta_{k}^l$ denotes the importance of edge type $k$, and is calculated by the following attention mechanism.
\begin{align}
	\begin{split}
		& w_{k}^l = \frac{1}{ \lvert \mathcal{V}_a \rvert} \sum\nolimits_{u \in \mathcal{V}_a}{\left(\bm{q}_a^l\right)^T \cdot \tanh \left(\bm{W}_a^l \left(\bm{h}^l_{u,k}\right)^T + \bm{b}_a^l \right)}\\
		& \beta_{k}^l = \text{softmax} \left(w_{k}^l\right)
	\end{split}
\end{align}
where $l = L-1, L-2,\ldots, 0$, denoting the order of neighborhood to the target node. $\bm{q}_a^l$ is a learnable node-type-specific semantic attention vector. $\bm{W}_a^l$ and $\bm{b}_a^l$ are node-type-specific learnable transformation matrix and bias vector, respectively.

For a node classification task, we apply a node-type-specific MLP layer to the final hidden layer as in \Cref{eqn:hetgtcn_down}.
\begin{equation}
	\label{eqn:hetgtcn_down}
	\bm{y}_u^\text{out} = \sigma \left(\bm{h}_u^0 \bm{W}_{\text{out},a} + \bm{b}_{\text{out},a} \right)
\end{equation}

\subsection{Heterogeneous Graph Tree Attention Network (HetGTAN)}
\label{HetGTAN}
The architecture of HetGTAN is the same as HetGTCN except that an edge-type-specific GTAN layer is applied to aggregate information of neighbors in the second module. Following the description in Section \ref{HetGTCN}, the proposed HetGTAN is formulated as below:
\begin{align}
	\label{eqn:HetGTAN}
	\begin{split}
		&\bm{h}_u^L = \bm{z}_u = \sigma \left( \bm{x}_u \bm{W}_{0,a} + \bm{b}_{0,a} \right) \\
		&\bm{h}_{u,k}^l = \text{ELU}\left( \sum\nolimits_{v\in \mathcal{N}_u^k} \alpha_{u,v,k}^{l+1} \bm{h}_v^{l+1} + \alpha_{u,u,k}^{l+1} \bm{z}_u \right)\\
		&\bm{h}_u^l = \sum_{k=1}^{K_a} {\beta_{k}^l \cdot \bm{h}_{u,k}^l}
	\end{split}
\end{align}
where $l = L-1, L-2,\ldots, 0$, denoting the order of neighborhood to the target node. $\alpha_{u,v,k}^l$ is the node-level attention weight for node pair $(u,v)$ through type $k$ connection, and $\beta_k$ is the attention weight for edge type $k$, which are calculated as below:
\begin{align}
	\begin{split}
		&\alpha_{u,v,k}^{l} =
		\begin{cases}
			\text{softmax} \left( \text{LeakyReLU} \left( \left[\bm{z}_u \parallel \bm{h}_v^{l}\right] \bm{a}_k^{l} \right)\right) & \text{if $u \neq v$}\\
			\text{softmax} \left(\text{LeakyReLU} \left( \left[\bm{z}_u \parallel \bm{z}_u\right] \bm{a}_k^{l} \right) \right) & \text{if $u = v$}
		\end{cases}\\
		& w_{k}^l = \frac{1}{ \lvert \mathcal{V}_a \rvert} \sum\nolimits_{u \in \mathcal{V}_a}{\left(\bm{q}_a^l\right)^T \cdot \tanh \left(\bm{W}_a^l \left(\bm{h}^l_{u,k}\right)^T + \bm{b}_a^l \right)}\\
		& \beta_{k}^l = \text{softmax} \left(w_{k}^l\right)\\
	\end{split}
\end{align}
where $\bm{a}_k^l$ is an edge-type-specific learnable attention vector.


\section{Experiments}
\label{exp}

In this section, we first evaluate the performance and efficiency of the proposed HetGTCN and HetGTAN on semi-supervised node classification tasks with three popular benchmark heterogeneous graphs in Section \ref{exp:performance}. We then demonstrate the deep capability of HetGTCN and HetGTAN in Section \ref{exp:deep}. Last, we compare the model performance with different aggregators in Section \ref{ablation}.

\subsection{Datasets}
\label{data}
We use two citation graphs ACM and DBLP, and a movie graph IMDB. The statistics of the three benchmark datasets are summarized in Table \ref{tab:dataset}. We use balanced label splits for both training and validation samples. The detailed descriptions are in Appendix \ref{appendix:data}. 


\begin{table*}[t]
	\caption{Statistics of benchmark datasets.}
	\label{tab:dataset}
	\centering
	\begin{tabular}{lllllccccc}
		\toprule
		Dataset &\#Node &\#Edge &\#Meta-Path &\#Node Feature &Target Node &\#Class &\#Training &\#Validation &\#Test \\
		\midrule
		\multirow{3}{*}{ACM} &Paper (P): 4,019 	&P-A: 13,407 &PAP: 57,853 	 &P: 1,902 &Paper &3 &600 &300 &3,119\\
		{} 					 &Author (A): 7,167 &P-S: 4,019  &PSP: 4,338,213 &A: 1,902 &{} &{} &{} &{} &{}\\
		{} 					 &Subject (S): 60 	&{} 		 &{}			 &S: 1,902 &{} &{} &{} &{} &{}\\
		\midrule
		\multirow{3}{*}{IMDB} &Movie (M): 4,278 &M-D: 4,278     &MDM: 17,446	&M: 3,066 &Movie &3 &300 &300 &3,678\\
		{} 					  &Director (D): 2,081 &M-A: 12,828 &MAM: 85,358 	&D: 3,066 &{} &{} &{} &{} &{}\\
		{} 					  &Actor (A): 5,257    &{} 			&{}				&A: 3,066 &{} &{} &{} &{} &{}\\
		\midrule
		\multirow{4}{*}{DBLP} &Author (A): 4,057 &A-P: 19,645 &APA: 11,113 &A: 334 &Author &4 &800 &400 &2,857\\
		{} 					  &Paper (P): 14,328 &P-T: 85,810 &APCPA: 5,000,495 &P: 334 &{} &{} &{} &{} &{}\\
		{} 					  &Term (T): 7,723 	 &P-C: 14,328 &APTPA: 7,043,571 &T: 50 &{} &{} &{} &{} &{}\\
		{} 					  &Conference (C): 20 &{} 		  &{} 				&C: 334 &{} &{} &{} &{} &{}\\
		\bottomrule
	\end{tabular}
\end{table*}

\subsection{Baselines} 
We compare the proposed HetGTCN and HetGTAN to nine HGNN baseline models including RGCN \cite{schlichtkrull2018modeling}, HAN \cite{wang2019heterogeneous}, MAGNN \cite{wang2019heterogeneous}, GTN \cite{yun2019graph}, HGT \cite{hu2020heterogeneous}, SimpleHGN \cite{lv2021we}, DMGI \cite{park2020unsupervised}, HetGCN and HetGAT. We also include the vanilla GCN \cite{kipf2017GCN} and GAT \cite{VelickovicCCRLB18} as baselines by ignoring the node and edge types. We have reviewed all baselines except HetGCN and HetGAT in Section \ref{literature}, and will not repeat the descriptions in this section.

In this section, we briefly introduce the two proposed baselines - HetGCN and HetGAT.

We apply the same architecture described in Section \ref{method} to turn the vanilla GCN \cite{kipf2017GCN} and GAT \cite{VelickovicCCRLB18} into HGNNs, and propose HetGCN and HetGAT accordingly in this section. Both models are consisted of three modules: (1) a node-type-specific transformation to project node features into the same vector space; (2) an edge-type-specific GCN or GAT layer to aggregate one-hop neighbors connected via the same type of edges; (3) a target-specific aggregator to combine representations of the target node from different edge types. We adopt the semantic attention aggregator in the third module for both HetGCN and HetGAT in this section. The detailed mathematical formulations of HetGCN and HetGAT are described in Appendix \ref{HetGCN_HetGAT}.

\subsection{Results on Node Classification}
\label{exp:performance}
The detailed experimental setups are described in Appendix \ref{exp:setup}. The experimental results of the proposed HetGTCN and HetGTAN and all baselines on the node classification task with all three benchmark datasets are summarized in Table \ref{tab:result_F1}. The evaluation metrics are Macro-F1 and Micro-F1. The results are obtained after 30 runs of each model. Mean and standard deviations are calculated after excluding the top and bottom $10\%$ data points. 
\begin{table*}[t]
	\caption{F1 score on node classification task, averaged by 30 runs of each model.}
	\label{tab:result_F1}
	\centering
		\begin{tabular}{ccccccc}
			\toprule
			{Dataset} &\multicolumn{2}{c}{ACM} &\multicolumn{2}{c}{IMDB} &\multicolumn{2}{c}{DBLP}\\
			{Metrics (\%)} &{Macro-F1} &{Micro-F1} &{Macro-F1} &{Micro-F1} &{Macro-F1} &{Micro-F1}\\
			\midrule
			RGCN &{91.6 $\pm$ 0.5} &{91.7 $\pm$ 0.4} &{58.0 $\pm$ 0.8} &{58.1 $\pm$ 0.6} &{92.8 $\pm$ 0.6} &{93.4 $\pm$ 0.5}\\
			HAN &{91.2 $\pm$ 0.2} &{91.0 $\pm$ 0.2} &{57.4 $\pm$ 0.4} &{57.2 $\pm$ 0.6} &{92.4 $\pm$ 0.2} &{93.2 $\pm$ 0.2}\\
			MAGNN &{90.7 $\pm$ 0.5} &{90.8 $\pm$ 0.5} &{58.7 $\pm$ 0.5} &{58.7 $\pm$ 0.7} &{93.0 $\pm$ 0.3} &{93.6 $\pm$ 0.2}\\
			GTN &{90.7 $\pm$ 0.3} &{90.1 $\pm$ 1.1} &{55.7 $\pm$ 0.8} &{55.4 $\pm$ 0.9} &{93.1 $\pm$ 0.4} &{94.1 $\pm$ 0.3}\\
			HGT &{87.1 $\pm$ 1.0} &{87.3 $\pm$ 1.4} &{56.3 $\pm$ 0.6} &{55.8 $\pm$ 0.9} &{91.6 $\pm$ 0.8} &{92.2 $\pm$ 0.6}\\
			SimpleHGN &{91.4 $\pm$ 0.4} &{91.5 $\pm$ 0.3} &{56.3 $\pm$ 0.5} &{56.1 $\pm$ 0.6} &{93.6 $\pm$ 0.2} &{94.3 $\pm$ 0.2}\\
			\midrule
			DMGI &{90.0 $\pm$ 0.6} &{90.5 $\pm$ 0.2} &{56.9 $\pm$ 1.0} &{56.3 $\pm$ 1.1} &{91.9 $\pm$ 0.3} &{92.9 $\pm$ 0.2}\\
			\midrule
			GCN &{90.7 $\pm$ 0.7} &{90.4 $\pm$ 1.0} &{57.1 $\pm$ 0.7} &{57.2 $\pm$ 0.9} &{83.5 $\pm$ 0.2} &{84.9 $\pm$ 0.3}\\
			GAT &{91.4 $\pm$ 0.6} &{91.2 $\pm$ 0.7} &{57.0 $\pm$ 1.1} &{56.3 $\pm$ 1.1} &{89.8 $\pm$ 0.8} &{90.9 $\pm$ 0.8}\\
			\midrule
			HetGCN &{91.1 $\pm$ 0.5} &{90.7 $\pm$ 0.9} &{57.8 $\pm$ 0.8} &{57.4 $\pm$ 0.8} &{92.1 $\pm$ 0.7} &{93.2 $\pm$ 0.5}\\
			HetGAT &{91.7 $\pm$ 0.4} &{91.7 $\pm$ 0.6} &{58.4 $\pm$ 0.9} &{57.8 $\pm$ 1.0} &{92.8 $\pm$ 0.5} &{93.6 $\pm$ 0.4}\\
			\midrule
			HetGTCN &{92.3 $\pm$ 0.2} &{92.1 $\pm$ 0.6} &{60.5 $\pm$ 0.5} &{60.0 $\pm$ 0.5} &{94.2 $\pm$ 0.2} &{94.8 $\pm$ 0.2}\\
			HetGTAN &{\textbf{92.3 $\pm$ 0.3}} &{\textbf{92.2} $\pm$ \textbf{0.4}} &{\textbf{60.8} $\pm$ \textbf{0.7}} &{\textbf{61.0} $\pm$ \textbf{0.5}} &{\textbf{94.4} $\pm$ \textbf{0.2}} &{\textbf{95.2} $\pm$ \textbf{0.2}}\\
			\bottomrule
		\end{tabular}
\end{table*}

It is shown that the proposed HetGTCN and HetGTAN consistently outperform all baselines on all three datasets, with 1-3$\%$ improvements over the second best model. HetGTAN generally yields better performance than HetGTCN, demonstrating that attention between each pair of nodes may better capture the heterogeneity in nodes and edges. This is also observed from the comparison of HetGCN and HetGAT. It is seen from Table \ref{tab:result_F1} that HetGAT outperforms HAN on all three datasets. It is not surprising because HAN manually selects meta-paths, which may lose information from other unselected meta-paths. For instance, in ACM dataset, a 2-hop HetGAT includes all 2-hop meta-paths: PAP, PSP, SPA, SPS, APA, APS, while HAN only selects two specific meta-paths: PAP and PSP. In addition, HAN ignores the intermediate nodes along the meta-paths, while HetGAT includes all intermediate nodes in the aggregation. In fact, HetGAT outperforms or matches most of the baselines, providing a new candidate baseline model for the future study.

We demonstrate the efficiency of the proposed HetGTCN and HetGTAN by comparing the time and memory consumption with all baselines. HAN, MAGNN and DMGI use 2-hop meta-paths for ACM and DBLP. For fair comparison, we use two graph layers for all other models including RGCN, GTN, HGT, SimpleHGN, HetGCN, HetGAT, HetGTCN, and HetGTAN, which is comparable to a 2-hop meta-path-based HGNN. The results are summarized in Table \ref{tab:efficency}.

It is seen from Table \ref{tab:efficency} that RGCN, HetGCN, HetGAT, HetGTCN, and HetGTAN consume comparable runtime, and are the most efficient models. All models consume similar memories, except the meta-path-based HGNNs including HAN, MAGNN, and GTN which consume significantly more memories on ACM and DBLP. The runtime and memory consumption of the meta-path-based HGNNs depend on the number of meta-path instances. Because ACM and DBLP have much more meta-path instances than IMDB, all three meta-path-based baselines consume significantly more runtime and memories than our proposed models on ACM and DBLP. The runtime of HAN is $\sim$2.7$\times$ that of HetGTCN on ACM, and $\sim$5$\times$ that of HetGTCN on DBLP. The runtime of GTN is $\sim$4.5$\times$ that of HetGTCN on ACM, and $\sim$8.4$\times$ that of HetGTCN on DBLP. MAGNN consumes the most runtime even we don't take the unscalable pre-processing time into consideration. The runtime of MAGNN is $\sim$44$\times$ that of HetGTCN on ACM, $\sim$11$\times$ that of HetGTCN on IMDB, and $\sim$717$\times$ that of HetGTCN on DBLP. HGT and SimpleHGN consume more runtime than HetGAT which may be due to the additional edge-type-specific learnable weight matrix and edge embeddings.


\begin{table*}[t]
	\caption{Time and memory consumption in ms/epoch and GB.}
	\label{tab:efficency}
	\centering
	\begin{threeparttable}
		\begin{tabular}{lllllll}
			\toprule
			\multirow{2}{*}{Model} &\multicolumn{2}{c}{ACM} &\multicolumn{2}{c}{IMDB} &\multicolumn{2}{c}{DBLP}\\
			\cmidrule{2-7}\\
			{} &{time (ms/epoch)} &{RAM (GB)} &{time (ms/epoch)} &{RAM (GB)} &{time (ms/epoch)} &{RAM (GB)}\\
			\midrule
			MAGNN &677.2 &10.6 &164.9 &2.3 &15,628.0\tnote{*} &1.6\tnote{*}\\
			DMGI &143.5 &1.5 &111.5 &1.4 &285.3 &1.8\\	
			SimpleHGN &72.9 &1.2 &72.2 &1.2 &113.0 &1.1\\
			GTN &70.3 &3.1 &23.6 &1.3 &183.3 &8.5\\
			HGT &44.6 &1.2 &43.7 &1.2 &96.7 &1.2\\
			HAN &42.0 &2.1 &11.0 &1.2 &109.5 &3.6\\
			RGCN &25.6 &1.1 &26.2 &1.2 &33.2 &1.2\\
			HetGAT &24.8 &1.2 &21.8 &1.3 &31.1 &1.2\\
			HetGTAN &18.5 &1.1 &17.0 &1.2 &27.4 &1.2\\
			HetGCN &15.8 &1.2 &15.5 &1.2 &22.0 &1.2\\
			HetGTCN &15.5 &1.1 &15.4 &1.2 &21.8 &1.2\\
			\bottomrule
		\end{tabular}
		\begin{tablenotes}
			\item[*] MAGNN runs on minibatch of DBLP with batch size of 8 samples.
		\end{tablenotes}
	\end{threeparttable}
\end{table*}

\subsection{Deep Capability}
\label{exp:deep}

It is sometimes desirable to increase the receptive field of an HGNN model so that it can capture information from high-order neighborhood. However, meta-path-based models suffer from scalability issue when going deep, due to the dramatically increasing number of meta-path instances. Regardless of the meta-path dependency, many HGNNs depend on GCN or GAT to aggregate information from edge-type-specific neighbors, therefore may suffer from the over-smoothing problem as GCN and GAT. The proposed HetGTCN and HetGTAN depend on GTCN and GTAN to aggregate information from edge-type-specific neighbors. The deep capability of GTCN and GTAN has been demonstrated in \cite{wu2022gtnet}.

\begin{figure*}[ht]
	\centering
	\begin{subfigure}[b]{0.3\textwidth}
		\centering
		\includegraphics[width = 1.0\linewidth]{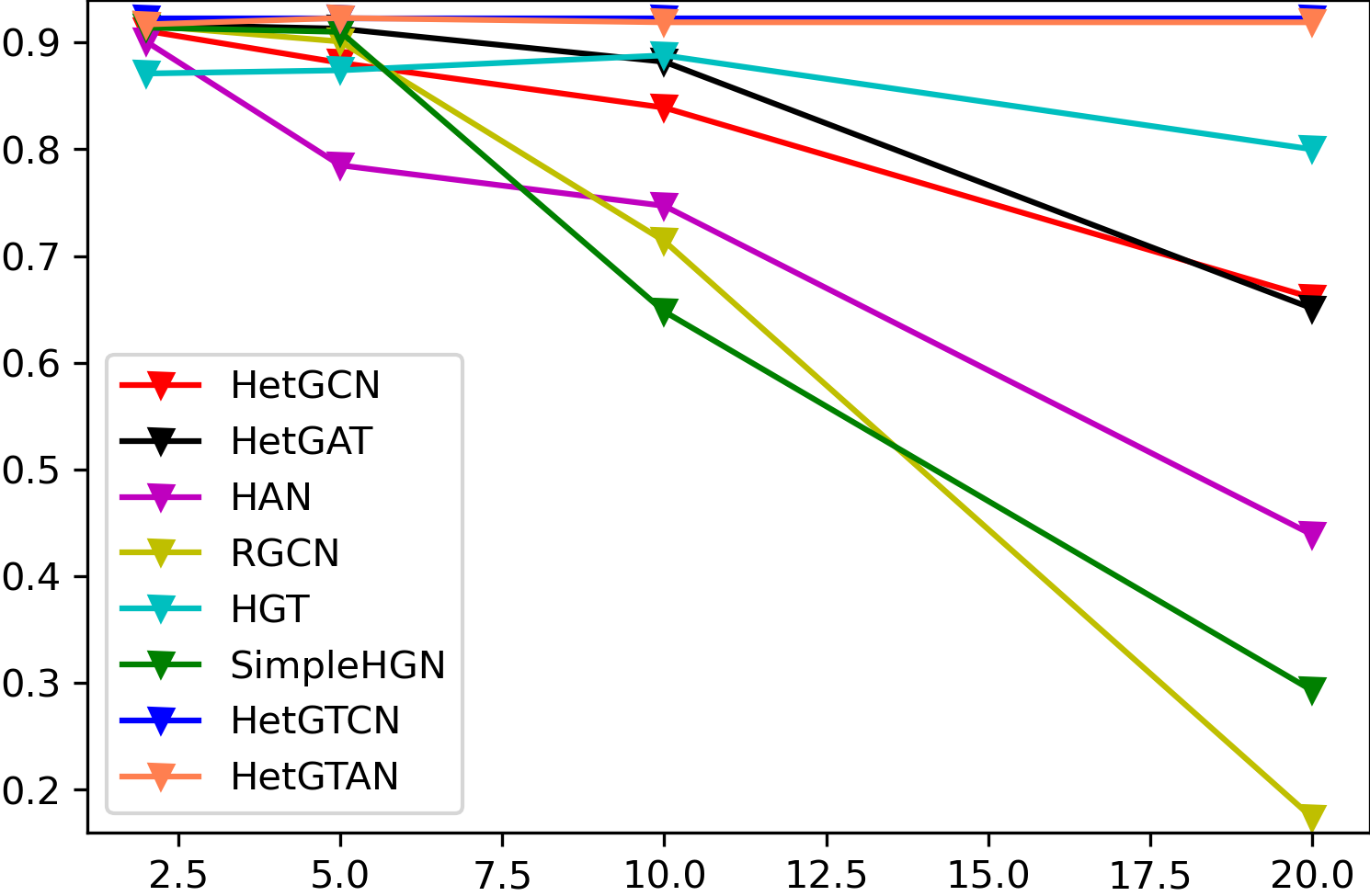}
		\caption{ACM}
		\label{fig:depth_ACM}
	\end{subfigure}
	\hfill
	\begin{subfigure}[b]{0.3\textwidth}
		\centering
		\includegraphics[width = 1.0\linewidth]{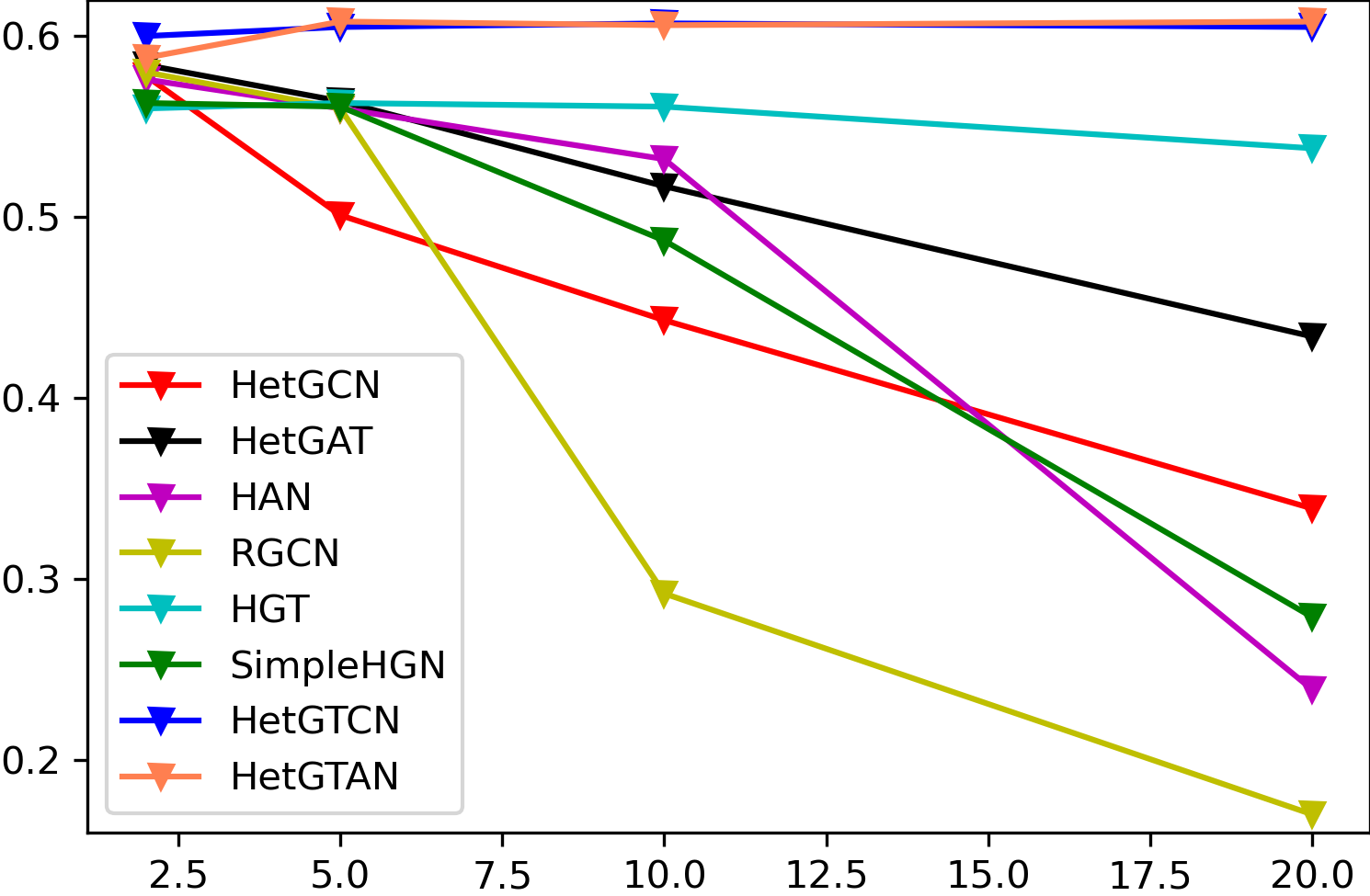}
		\caption{IMDB}
		\label{fig:depth_IMDB}
	\end{subfigure}
	\hfill
	\begin{subfigure}[b]{0.3\textwidth}
		\centering
		\includegraphics[width = 1.0\linewidth]{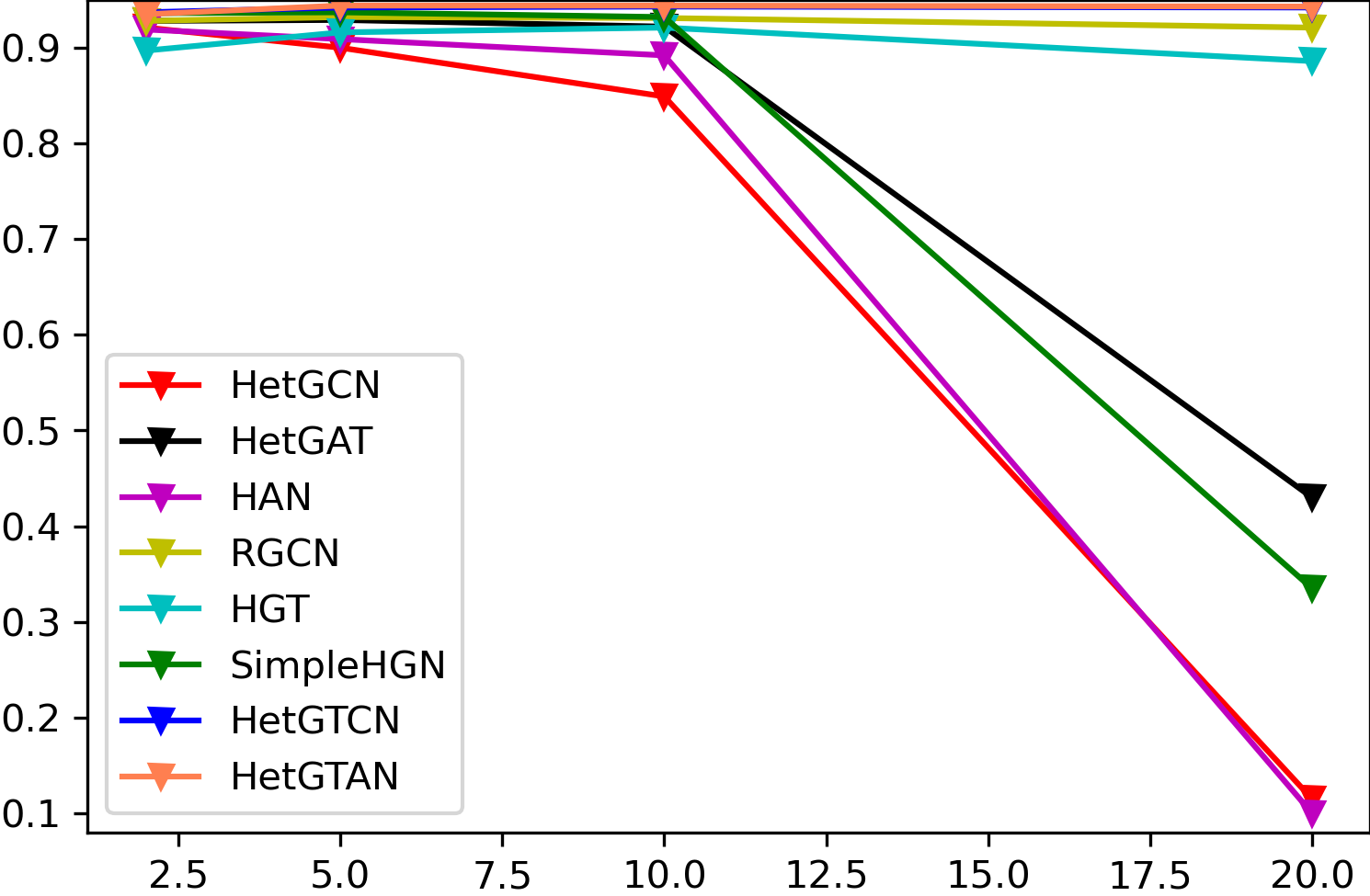}
		\caption{DBLP}
		\label{fig:depth_DBLP}
	\end{subfigure}
	\caption{Model performance in terms of Macro-F1 score at different depths.}
	\label{fig:depth}
\end{figure*}

In this section, we compare the performance of the proposed HetGTCN and HetGTAN with six baseline HGNNs including RGCN, HAN, HGT, SimpleHGN, HetGCN, and HetGAT with different number of message passing layers (2, 5, 10, and 20). We do not include MAGNN, GTN, and DMGI in this comparison due to the runtime and memory issue with these models. The results are displayed in \Cref{fig:depth}.

\textbf{Results}. It is seen from Figure \ref{fig:depth} that the performance of the proposed HetGTCN and HetGTAN does not compromise when going deep while the performance of all other baselines degrades significantly with an increasing model depth. Among all baselines, the performance of HGT degrades much slower than other baselines, which may be due to the skip connection.

\subsection{Ablation Study}
\label{ablation}
We adopt the semantic attention aggregator in HetGTCN and HetGTAN to combine edge-type-specific representations in previous sections. In this section, we conduct an ablation study to explore different aggregators in HetGTCN and HetGTAN. HetGTCN\textsubscript{mean} and HetGTAN\textsubscript{mean} adopt a simple mean aggregator in the third module to combine edge-type-specific representations, where $\bm{h}_u^l = \frac{1}{K_a} \sum_{k=1}^{K_a} \bm{h}_{u,k}^l$. HetGTCN\textsubscript{LW} and HetGTAN\textsubscript{LW} adopt a weighted sum aggregator in the third module, where $\bm{h}_u^l = \sum_{k=1}^{K_a} \theta_k^l \bm{h}_{u,k}^l$, and the weight $\theta_k^l$ for each edge type is directly learned. For HetGTAN, we have an additional variant HetGTAN\textsubscript{ns} which removes the semantic attention layer, and the message passing rule in the second module becomes
\begin{equation}
	\bm{h}_{u}^{l} = \text{ELU}\left(\sum\nolimits_{k=1}^{K_a} \left(\sum\nolimits_{v\in \mathcal{N}_u^k} \alpha_{u,v,k}^{l+1} \bm{h}_v^{l+1} + \alpha_{u,u,k}^{l+1} \bm{z}_u \right)\right)
\end{equation}

Other settings of these variants stay the same as the original models. The average Macro-F1 score on the node classification task with all three datasets are summarized in Table \ref{tab:ablation}. The simple mean and semantic attention aggregator yield comparable performance for HetGTCN on ACM and IMDB, while HetGTCN with the semantic attention aggregator works best on DBLP. The simple mean, weighted sum, and semantic attention aggregator all yield comparable performance for HetGTAN on IMDB and DBLP, while HetGTAN with the semantic attention aggregator works best on ACM. Overall, the semantic attention aggregator works best for both HetGTCN and HetGTAN on all three datasets.

\begin{table}[t]
	\caption{Ablation study for HetGTCN and HetGTAN, using 5 layers for all models.}
	\label{tab:ablation}
	\begin{small}
		\begin{tabular}{cccc}
			\toprule
			{Dataset} &{ACM} &{IMDB} &{DBLP}\\
			{Metrics (\%)} &{Macro-F1} &{Macro-F1} &{Macro-F1}\\
			\midrule
			HetGTCN\textsubscript{mean} &92.3 $\pm$ 0.2 &60.5 $\pm$ 0.7 &91.2 $\pm$ 0.7\\
			HetGTCN\textsubscript{LW} &91.7 $\pm$ 1.3 &60.1 $\pm$ 0.7 &93.8 $\pm$ 0.7\\
			HetGTCN &92.3 $\pm$ 0.2 &60.5 $\pm$ 0.5 &94.2 $\pm$ 0.2 \\
			\midrule
			HetGTAN\textsubscript{ns} &91.8 $\pm$ 0.3 &60.4 $\pm$ 0.5 &94.1 $\pm$ 0.3\\
			HetGTAN\textsubscript{mean} &91.9 $\pm$ 0.3 &60.9 $\pm$ 0.7 &94.4 $\pm$ 0.2\\
			HetGTAN\textsubscript{LW} &91.9  $\pm$ 0.5 &60.9 $\pm$ 0.5 &94.4 $\pm$ 0.2\\
			HetGTAN &92.3 $\pm$ 0.3 &60.8 $\pm$ 0.7 &94.4 $\pm$ 0.2\\
			\bottomrule
		\end{tabular}
	\end{small}
\end{table}

\section{Conclusion}
In this paper, we propose two heterogeneous graph neural network models - HetGTCN and HetGTAN which are based on Graph Tree Networks. Both models are meta-path-free, and are able to go deep without compromising performance. We conduct comprehensive experiments on three real-world heterogeneous graph data with unified pre-processing and demonstrate the effectiveness and efficiency of our proposed models. We also propose two baseline HGNN models - HetGCN and HetGAT to extend the vanilla GCN and GAT models to heterogeneous graphs. Future work could include exploring potentials of the proposed HetGTCN and HetGTAN on various interesting heterogeneous graph datasets for different downstream tasks.


\bibliographystyle{IEEEtran}
\bibliography{IEEEabrv, HetGTree}

\appendices
\section{Datasets}
\label{appendix:data}
\textbf{ACM}\footnote{https://dl.acm.org/} is a citation graph. It covers papers published in KDD, SIGMOD, VLDB, SIGCOMM and MOBICOMM. The papers are divided into three classes: Data Mining, Database, and Wireless Communication. The task is to classify the type of unlabeled papers. There are three types of nodes: Paper, Author and Subject, and four types of directed edges: Paper-Author (PA), Paper-Subject (PS) and their reverse connections. The feature of a paper node is obtained from the bag-of-words representations of its keywords. The feature of an author is obtained by averaging the features of papers that the author is connected with. The feature of a subject is calculated by averaging the features of papers that the subject is connected with. We use the same data processing as in GTN \cite{yun2019graph} and MAGNN \cite{fu2020magnn} when a node's feature is not available. We choose 600 balanced samples as the training set, another 300 balanced samples as the validation set, and all remaining 3,119 samples as the test set.

\textbf{IMDB}\footnote{https://www.imdb.com/} has three types of nodes: Movie, Director and Actor, and four types of directed edges: Movie-Director (MD), Movie-Actor (MA) and their reverse connections. The task is to predict the types of movies in one of the three classes: Action, Drama and Comedy. We use a subset of IMDB provided by the Pytorch Geometric library \cite{Fey/Lenssen/2019}, which is the same subset as used by MAGNN \cite{fu2020magnn}. The feature of a movie is obtained from the bag-of-words representations of its plot keywords. The features of a director and an actor are obtained by averaging the features of movies that are connected with the director and actor, respectively. We choose 300 balanced samples as the training set, another 300 balanced samples as the validation set, and all remaining 3,678 samples as the test set.

\textbf{DBLP}\footnote{https://dblp.uni-trier.de/} is a computer science bibliography website. It contains four types of nodes: Author, Paper, Term, and Conference, and six types of directed edges: Author-Paper (AP), Paper-Term (PT), Paper-Conference (PC) and their reverse connections. The task is to classify the authors into one of the four research areas: database, data mining, artificial intelligence, and information retrieval. We use a subset of DBLP provided by the Pytorch Geometric library \cite{Fey/Lenssen/2019}, which is the same subset as used by MAGNN \cite{fu2020magnn}. The feature of an author is obtained from the bag-of-words representations of the author's papers. The feature of a paper is obtained by averaging the features of authors that the paper is connected with. The feature of a conference is obtained by averaging of the features of papers that the conference is connected with. The features of term nodes are provided by the DBLP dataset. We choose 800 balanced samples as the training set, another 400 balanced samples as the validation set, and all remaining 2,857 samples as the test set.

\section{Proposed Baselines - HetGCN and HetGAT}
\label{HetGCN_HetGAT}
\textbf{HetGCN.} The mathematical formulation of HetGCN is described as below:
\begin{align}
	\label{eqn:HetGCN}
	\begin{split}
		&\bm{h}_u^0 = \bm{z}_u = \sigma \left( \bm{x}_u \bm{W}_{0,a} + \bm{b}_{0,a} \right) \\
		&\bm{h}_{u,k}^{n} = \sum\nolimits_{v \in {\mathcal{N}_u^k} \cup \{u\}}{\hat{\bm{A}}_{k,uv} \bm{h}^{n-1}_v  \bm{W}^{n-1}  }  \\
		&\bm{h}_u^n = \text{ReLU} \left(\sum_{k=1}^{K_a} {\beta_{k}^n \cdot \bm{h}_{u,k}^n}\right)
	\end{split}
\end{align}
where $n = 1, 2, \ldots, L$, denoting the message passing layer number. $\beta_k^n$ is the semantic attention weight for the edge type $k$, and is calculated as
\begin{align}
	\begin{split}
		& w_{k}^{n} = \frac{1}{ \lvert \mathcal{V}_a \rvert} \sum\nolimits_{u \in \mathcal{V}_a}{\left(\bm{q}_a^{n}\right)^T \cdot \tanh \left(\bm{W}_a^{n} \left(\bm{h}^{n}_{u,k}\right)^T + \bm{b}_a^{n} \right)}\\
		& \beta_{k}^n = \text{softmax} \left(w_{k}^n\right)
	\end{split}
\end{align}

\textbf{HetGAT.} The mathematical formulation of HetGAT is described as below:
\begin{align}
	\label{eqn:HetGAT}
	\begin{split}
		&\bm{h}_u^0 = \bm{z}_u = \sigma \left( \bm{x}_u \bm{W}_{0,a} + \bm{b}_{0,a} \right) \\
		&\bm{h}_{u,k}^{n} = \text{ELU} \left(
		\sum\nolimits_{v \in {\mathcal{N}_u^k} \cup \{u\}} {\alpha_{u,v,k}^{n-1} \bm{h}_v^{n-1} \bm{W}^{n-1}}  \right) \\
		&\bm{h}_u^n = \text{ReLU} \left(\sum_{k=1}^{K_a} {\beta_{k}^n \cdot \bm{h}_{u,k}^n}\right)\\
	\end{split}
\end{align}
where $n = 1, 2, \ldots, L$, denoting the message passing layer number. $\alpha_{u,v,k}^n$ is the node-level attention weight for node pair $(u,v)$ through type $k$ connection, and $\beta_k^n$ is the semantic attention weight for edge type $k$, which are calculated as
\begin{equation}
	\begin{split}
		&\alpha_{u,v,k}^{n} = \text{softmax}\left(
		\text{LeakyReLU}\left( \left[\bm{h}_u^{n} \bm{W}^{n} \parallel \bm{h}_v^{n} \bm{W}^{n} \right] \bm{a}_k^{n} \right) \right)\\
		& w_{k}^{n} = \frac{1}{ \lvert \mathcal{V}_a \rvert} \sum\nolimits_{u \in \mathcal{V}_a}{\left(\bm{q}_a^{n}\right)^T \cdot \tanh \left(\bm{W}_a^{n} \left(\bm{h}^{n}_{u,k}\right)^T + \bm{b}_a^{n} \right)}\\
		& \beta_{k}^n = \text{softmax} \left(w_{k}^n\right)
	\end{split}
\end{equation}

\section{Experimental Setup}
\label{exp:setup}
We use 64 hidden units for all models, except that we use 128 hidden units for the semantic attention layer in HAN, HetGCN, HetGAT, HetGTCN and HetGTAN. DMGI is trained using a maximum epoch of 10,000 and an early stopping patience of 20. MAGNN is trained using a maximum epoch of 100 and an early stopping patience of 10. All other models are trained using a maximum epoch of 500 and an early stopping patience of 100.

\textbf{GCN} \cite{kipf2017GCN}. We use a two-layer model. The dropout rate is set to 0.5 for all datasets, and is applied to the initial projection layer and the output of each intermediate GCN layer. Adam optimizer is used with a learning rate of 0.005 and weight decay of 5e-5 for all three datasets.

\textbf{GAT} \cite{VelickovicCCRLB18}. We use a two-layer model with one attention head. There are two dropouts: one is the dropout after each intermediate layer and another is the attention dropout. The dropout rates are (0.8, 0.2), (0.8, 0.2) and (0, 0) for ACM, IMDB and DBLP, respectively. Adam optimizer is used with a learning rate of 0.005 and weight decay of 5e-5 for all three datasets.

\textbf{RGCN} \cite{schlichtkrull2018modeling}. We use a two-layer model. The dropout rates are set to 0.5, 0.5 and 0 for ACM, IMDB and DBLP, respectively. The number of bases is set to 5. Adam optimizer is used with a learning rate of 0.005 and weight decay of 1e-5 for all three datasets. Since RGCN is designed for multi-relational knowledge graphs with a single node type, we first apply a node-type-specific transformation to project all node features into the same vector space, and then implement RGCN with the code provided by the Pytorch Geometric library \cite{Fey/Lenssen/2019}.

\textbf{HAN} \cite{wang2019heterogeneous}. We use the same settings as the original paper \cite{wang2019heterogeneous}. The pre-selected meta-paths of ACM are PAP and PSP. The pre-selected meta-paths of IMDB are MDM and MAM. The pre-selected meta-paths of DBLP are APA, APCPA and APTPA. The number of attention heads is 8, and the dropout rate is 0.6. Adam optimizer is used with a learning rate of 0.005 and weight decay of 0.001 for all three datasets. We implement HAN with the code provided by DGL \cite{wang2019dgl}.

\textbf{MAGNN} \cite{fu2020magnn}. All settings are the same as the original paper except that we use 64 hidden units. The pre-selected meta-paths are the same as used by HAN \cite{wang2019heterogeneous} for all three datasets. The number of attention heads is 8 with each head having 8 hidden units. The dropout rate is 0.5. Adam optimizer is used with a learning rate of 0.005 and weight decay of 0.001 for all three datasets. \cite{fu2020magnn} runs on full batch of ACM and IMDB, and minibatch\footnote{Running on full batch of DBLP results in the Out-of-Memory issue.} of DBLP with a batch size of 8 and the number of neighbor samples as 100. The preprocessing time of MAGNN is at least O($N^3$) such that we are unable to extract the meta-path-based neighbors for DBLP due to the Out-of-Memory and Out-of-Time issue. Therefore, we use the preprocessed meta-path information provided by the authors, and implement MAGNN with the authors' official code\footnote{https://github.com/cynricfu/MAGNN}.

\textbf{GTN} \cite{yun2019graph}. For ACM and IMDB, we use 2 channels and 32 hidden units for each channel. For DBLP, we use 2 channels and 16 hidden units for each channel as using 32 hidden units for each channel results in the Out-of-Memory issue with our 12 GB vRAM RTX 3060 GPU. We use two GTN layers for ACM and DBLP, and three GTN layers for IMDB. Adam optimizer is used with a learning rate of 0.005 and weight decay of 0.001 for all three datasets. We implement GTN with the authors' official code\footnote{https://github.com/seongjunyun/Graph\_Transformer\_Networks}.

\textbf{HGT} \cite{hu2020heterogeneous}. The hyperparameters are fine-tuned for all three datasets to achieve the best performance. The number of attention heads is set to 4. We use a two-layer model for ACM and IMDB, and a three-layer model for DBLP. The dropout rate is set to 0.2. Adam optimizer is used with a learning rate of 0.005 and weight decay of 5e-5 for all three datasets. We implement HGT with the code provided by the Pytorch Geometric library \cite{Fey/Lenssen/2019}.

\textbf{SimpleHGN} \cite{lv2021we}. The feature and attention dropout rates are both set to 0.5 for all three datasets. $\beta$ is set to 0.05. Adam optimizer is used with a learning rate of 0.01 and weight decay of 0 for all three datasets. We use a two-layer model with eight attention heads for all three datasets which yield matching or better results than with the settings in the original paper \cite{lv2021we}. We implement SimpleHGN with the authors' official code\footnote{https://github.com/thudm/hgb}.

\textbf{DMGI} \cite{park2020unsupervised}. We use the same settings as the original paper \cite{park2020unsupervised}. The authors use the same pre-selected meta-paths as HAN \cite{wang2019heterogeneous} does. The dropout rate is 0.5. Adam optimizer is used with a learning rate of 0.0005 and weight decay of 0.0001 for all three datasets. The self-connection weight is set to 3. The consensus regularization coefficient $\alpha$ is set to 0.001 and the semi-supervised loss coefficient $\gamma$ is set to 0.1. We implement DMGI with the authors' official code\footnote{https://github.com/pcy1302/DMGI}.

\textbf{HetGCN.} We use a two-layer model. The dropout rate is set to 0.5, 0.5 and 0 for ACM, IMDB and DBLP, respectively. Adam optimizer is used with a learning rate of 0.005 and weight decay of 1e-5 for all three datasets.

\textbf{HetGAT.} We use a two-layer model with one attention head. There are two dropouts: one is the dropout for the feature projection layer, and the other is the dropout after each intermediate layer. The dropout rates are set to (0.8, 0.2), (0.8, 0.2) and (0, 0) for ACM, IMDB and DBLP, respectively. Adam optimizer is used with a learning rate of 0.005 and weight decay of 5e-5 for all three datasets.

\textbf{HetGTCN.} We use a five-layer model. There are also two dropouts: one is the dropout for the feature projection layer, and the other is the dropout after each intermediate layer. The dropout rates are set to (0.8, 0.6) for all three datasets. Adam optimizer is used with a learning rate of 0.005 and weight decay of 1e-5 for all three datasets.

\textbf{HetGTAN.} We use a five-layer model. Other settings are the same as HetGAT.

The code to replicate our experiments is available at \url{https://github.com/hetgnn/hetGTNet}.


\end{document}